\documentclass[final]{cvpr}

\usepackage{times}
\usepackage{epsfig}
\usepackage{graphicx}
\usepackage{amsmath}
\usepackage{amssymb}
\usepackage{booktabs}

\usepackage[pagebackref=true,breaklinks=true,colorlinks,bookmarks=false]{hyperref}

\def\cvprPaperID{****} 
\def\confYear{CVPR 2021}

\begin{document}

\title{Team RUC AI$\cdot$M$^3$ Technical Report at ActivityNet 2021:\\ Entities Object Localization}

\author{Ludan Ruan\textsuperscript{1}, Jieting Chen\textsuperscript{1}, Yuqing Song\textsuperscript{1}, Shizhe Chen\textsuperscript{2}, Qin Jin\textsuperscript{1\thanks{Corresponding author.}} \\
	\textsuperscript{1}School of Information, Renmin University of China \textsuperscript{2}INRIA \\
	\tt\small \{ruanld,jietingchen,syuqing,qjin\}@ruc.edu.cn, cshizhe@gmail.com}

\maketitle

\begin{abstract}
Entities Object Localization (EOL) aims to evaluate how grounded or faithful a description is, which consists of caption generation and object grounding. 
Previous works tackle this problem by jointly training the two modules in a framework, which limits the complexity of each module.
Therefore, in this work, we propose to divide these two modules into two stages and improve them respectively to boost the whole system performance.
For the caption generation, we propose a Unified Multi-modal Pre-training Model (UMPM) to generate event descriptions with rich objects for better localization.
For the object grounding, we fine-tune the state-of-the-art detection model MDETR and design a post processing method to make the grounding results more faithful. 
Our overall system achieves the state-of-the-art performances on both sub-tasks in Entities Object Localization challenge at Activitynet 2021, with 72.57 localization accuracy on the testing set of sub-task I and 0.2477 F1\_all\_per\_sent on the hidden testing set of sub-task II.
\end{abstract}

\section{Introduction}
\begin{figure}[h]
\centering
\includegraphics[width=0.45\textwidth]{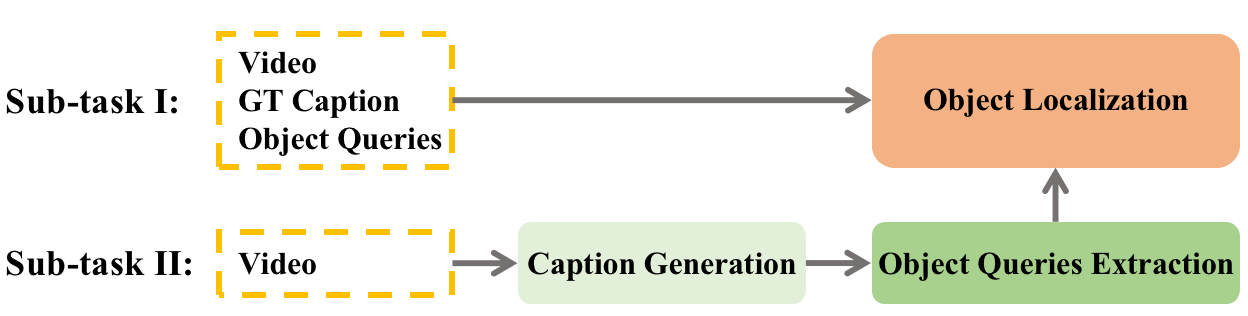}
\setlength{\abovecaptionskip}{10pt}
 \caption{Illustration of the two sub-tasks in the EOL Challenge.}
\label{fig:sub-tasks}
\end{figure}

The goal of the \textit{Entities Object Localization (EOL)} task~\cite{Zhou_GVD_CVPR19} is to evaluate how grounded or faithful a description (generated or annotated) is to the video.
As shown in Figure~\ref{fig:sub-tasks}, there are two sub-tasks in this challenge: grounding with Ground-Truth (GT) sentences (sub-task I) and grounding with generated sentences (sub-task II).
For the sub-task I, the EOL task is simplified to grounding the given object words with spatial bounding boxes in the video frames.
For the sub-task II, it is more challenging and consists of two stages: 1) automatically generating descriptions for the video, and 2) identifying the object words in the description and localizing them in the video.
The system prediction is compared against the human annotation to determine the correctness and overall localization accuracy. 

Previous works~\cite{Zhou_GVD_CVPR19} tackle this problem by jointly training the caption generation module and the object grounding module in one model, which however limits the complexity of each module.
To keep the advantages of each module and better serve the two sub-tasks, we propose to train the two modules independently and solve this problem in a two-stage manner.
To generate video descriptions with rich objects for better localization, we propose a concept enhanced Unified Multi-modal Pre-training Model~(UMPM) based on multi-modal pre-training framework, which is also used in the Dense Video Captioning task of the ActivityNet 2021 Challenge.
To correctly localize the given object words, we fine-tune an effective text-conditioned detector MDETR~\cite{Kamath_MDETR_arxiv2021} on the Activityet Entities dataset and further propose a post processing method to adapt to this task.
Specifically, for the sub-task I, we localize the given object words in the ground-truth captions with the fine-tuned MDTER, and for the sub-task II, we first parse the object words in the descriptions generated by our UMPM module and then localize them with the MDTER.

Experimental results on the ActivityNet Entities dataset demonstrates the effectiveness of our two-stage method.
Our overall system achieves the state-of-the-art results on both sub-tasks, with 72.57 localization accuracy on the   testing set of sub-task I and 0.2477 F1\_all\_per\_sent on the hidden testing set of sub-task II.

The remainder of this paper is organized as follows: Section 2 introduces the overall framework of our Entities Object Localization System, including the caption generation module and the object grounding module.
Section 3 presents the experimental results and analysis of the two sub-tasks. 
Finally, Section 4 concludes the paper.

\begin{figure*}[h]
\centering
\includegraphics[width=0.95\textwidth]{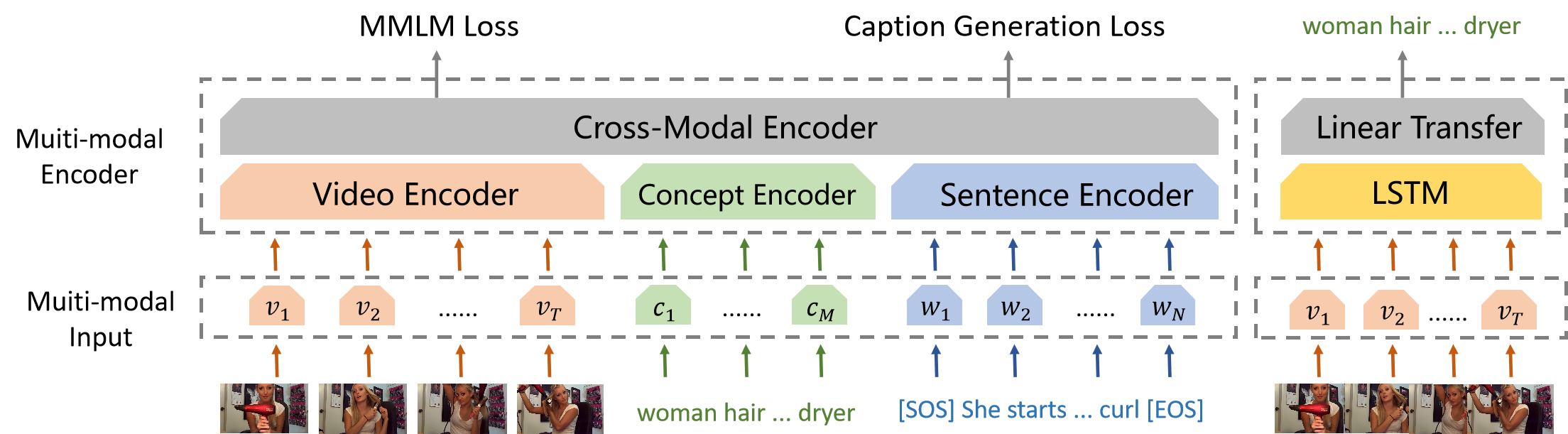}
 \caption{Illustration of the proposed Unified Multi-modal Pre-training Model and Concept Predictor.}
\label{fig:UMPM}
\end{figure*}

\section{Entities Object Localization System}
In this section, we introduce our entities object localization system, which consists of two major modules: 1) the Caption Generation Module and 2) the Object Localization Module.

\subsection{Caption Generation Module}
Motivated by the success of transformer-based multi-modal pre-training models~\cite{Li_Oscar_ECCV20} on many V+L downstream tasks, we propose to generate event captions via a Unified Multi-modal Pre-training Model (UMPM) as shown in Figure~\ref{fig:UMPM}. 
To generate more detailed descriptions that contain rich objects for better localization, we further enhance the model with semantic concepts, which is predicted by another concept predictor.
To be specific, we introduce our caption generation module in details with respect to the following components: 1) Multi-modal Input; 2) Model Architecture; 3) Pre-training Objective and 4) Fine-tuning for Video Caption Generation.

\paragraph{Multi-modal Input}
The input sequence to our UMPM model consists of three parts, including video sequence, concept labels and caption sentence.
Given an untrimmed video, we divide it into non-overlapping clips with 64 frames per segment. Then we extract clip level features as in our previous work~\cite{Song_anett2_2020, Chen_anett3_2019}, which includes: 1) Resnet200~\cite{He_resnet_CVPR16} from the image modality pretrained on the ImageNet dataset; 2) I3D~\cite{Carreira_i3d_CVPR17} from the motion
modality pretrained on the Kinetics dataset;
These two features are temporally aligned and are concatenated together as the feature for the $t_{th}$ segment. Therefore, we get the video clip feature sequence $V = \{v_1, v_2, ... ,v_T\}$.

Then, we automatically predict semantic concept labels based on the video features.
We pre-train a LSTM-based concept predictor with multi-label classification loss, where the ground-truth concept labels are the nouns and verbs extracted from the annotated captions.
After pre-training, we average the predicted probabilities of each frame and choose the top $M$ predicted concepts to input to the captioning model.

For the captioning sentence, we add a special start token ([SOS]) and a special end token ([EOS]) to the start and end of the sentence, and represent each token in the sentence with a word embedding vector learned from scratch.





\paragraph{Model Architecture}
Our UMPM contains four multi-layer transformer encoders, including three independent encoders and a cross encoder. 
We first employ independent encoders to encode the video sequence, concept labels and captioning sentence respectively to capture their intra-context information. 
The outputs of the three encoders are then concatenated as a whole input sequence to the cross encoder to encode the inter-context across different modalities. 

\paragraph{Pre-training Objective}
Similar to the V+L pre-training models~\cite{chen2019uniter,li2020unicoder,lu2019vilbert}, we adopt the multi-modal mask language modeling (MMLM) as the pre-training task.
We randomly choose 15\% word tokens for prediction.
Each chosen word is replaced with a special [MASK] token 80\% of the time, another random word 10\% of the time and the original word 10\% of the time as in~\cite{devlin2019bert}.
The fused feature from cross encoder is fed to an output softmax layer which is tied with the input embedding to predict the original word.
The training objective of MMLM can be expressed as follows:
\begin{equation}
    \mathcal{L}_{MMLM} = - \mathbb{E}_{(V,S)\sim\mathcal{D}} \log p(w_m|w_{\setminus m},V,C;\Theta)
\end{equation}
where $\mathcal{D}$ denotes the whole training set, $w_m$ denotes the masked words in $S$, $C$ denotes the predicted concept labels from $V$, and $\Theta$ denotes all learnable parameters of the pre-training model.
With the MMLM pre-training task, the model can learn the semantic alignments between video and sentence, and also learn the language structure of sentences.


\begin{figure*}[htbp]
\centering
\includegraphics[width=0.95\textwidth]{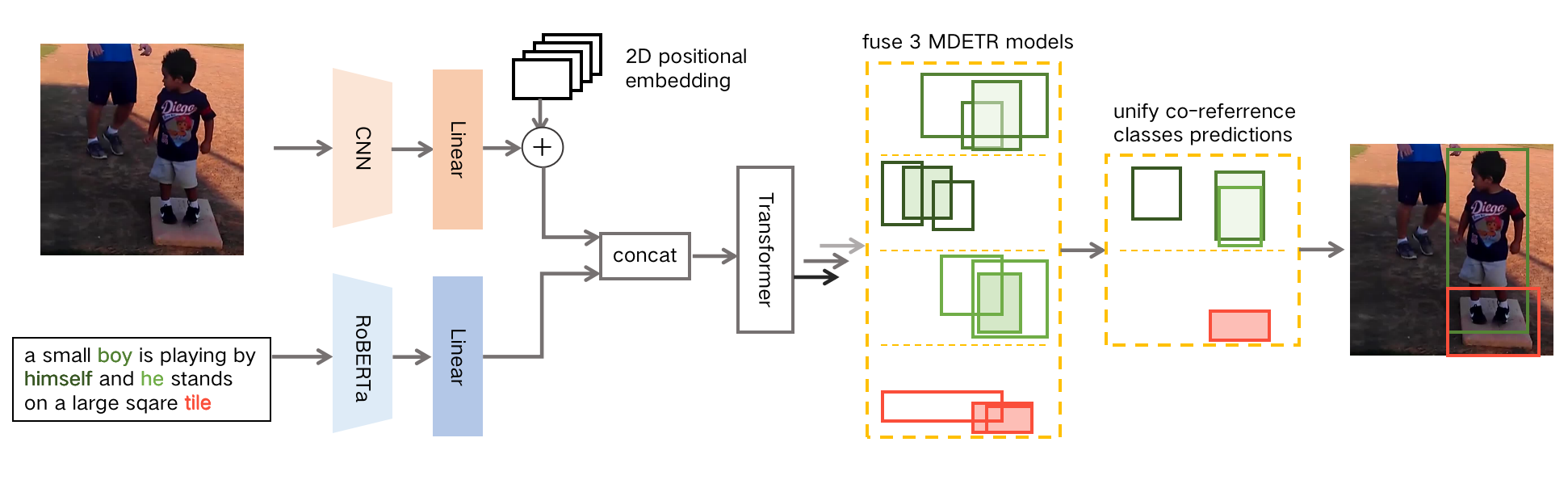}
 \caption{Illustration of the Object Localization module.}
\label{fig:mdetr_structure}
\end{figure*}

\paragraph{Fine-tuning for Video Caption Generation}
To enable caption generation, we finetune the pre-trained UMPM model using seq2seq objective. We randomly mask out 15\% of the caption tokens and use the corresponding output representations to perform classification to predict the token ids. Note that self-attention mask is constrained on caption tokens after the generated position in all transformer layers. 

During inference, we first encode the video clip sequence, concept labels, and the start token [SOS] as inputs. Then the model starts generation by feeding in a [MASK] token in the first caption token position. After predicting the token id of the [MASK], the first caption token is replaced with the predicted token and a new [MASK] is appended for the next word prediction. The process continues until the model outputs the [EOS] token.

\subsection{Object Localization Module}
Our object localization module is based on MDETR~\cite{Kamath_MDETR_arxiv2021}, which is the state-of-the-art text-conditioned object detection model for image.
We fine-tune the model on the ActivityNet Entities dataset and propose a post processing approach named IoU voting for better localization results.
For the sub-task I, we use the ground-truth object words as queries, while for the sub-task II, we extract object queries from the generated captions.
We introduce our object localization module in details with respect to the following components: 1) Object Queries Extraction, 2) Model Architecture and 3) Post Processing.

\paragraph{Object Queries Extraction}
To extract object queries from generated sentences for sub-task II, we first use Spacy to parse the sentence and then select all the nouns and pronouns in the generated descriptions. The words corresponding to the official object vocabulary are the candidates to be localized. Furthermore, we find that there are some abstract nouns which do not appear in the video. For example, in the caption of ``A woman is seen speaking to the camera while holding a accordion'', the word ``camera'' means shot but not real camera and does not appear in the video. Therefore, we filter out these abstract words for simplicity.




\paragraph{Model Architecture}
As shown in Figure~\ref{fig:mdetr_structure}, following MDETR~\cite{Kamath_MDETR_arxiv2021}, the object localization module consists of two main parts: the multi-modal encoders for the text and image encoding, and a transformer model for the detection.
For the multi-modal encoding, a CNN such as EfficientNet~\cite{tan2019efficientnet} is used to extract image features, and a pre-trained language model such as RoBERTa~\cite{liu2019roberta} is used to extract text features.
Then, these two features are mapped into a joint embedding space, and then concatenated to fed to the transformer detection module which is based on DETR~\cite{carion2020end}. 
The transformer detection module consists of a cross encoder and a decoder, which takes the fused feature as input and outputs box candidates, including the coordinate and its soft token prediction.
The soft token prediction is the matching score of each token span in the text.

The MDETR is pre-trained by two objectives for better alignment between image and text, including soft token prediction and contrastive alignment.
Instead of predicting a categorical class from the fixed vocabulary dictionary for each box candidate, MDETR regulates the number of tokens for each text into 256, then predicts the matching scores of token spans with the bounding box. In this way, we can localize an region for a span of tokens in free form text. It also benefits the long-tail distribution dataset.
The contrastive alignment loss will pull the embedding of visual object and its corresponding text token closer compared to unrelated tokens in the joint space, which is good for text-conditioned object detection.

\paragraph{Post Processing}
To reduce prediction noise and further improve the performance, we use different random seeds to train multiple models and get several box candidates for each object query. We design a method called IoU voting to fuse the results. 
For each box candidate, we calculate its IoUs with other box candidates and sum them up as a confident score. Then we pick the box with the highest score.

Besides, in the ActivityNet Entities Dataset, several classes in a sentence may refer to the same object in the image, which are called co-referrence classes. However, MDETR generates the box prediction for each of them. Thus, we use IoU voting again to unify these candidates.



\section{Experiments}

\subsection{Dataset}
We carry out experiments on the ActivityNet-Entities dataset~\cite{Zhou_GVD_CVPR19} for Entities Object Localization. 
The whole dataset contains 20K videos with 158k noun phrase (NP) grounded bounding box annotations.
We follow the official split with 10k videos for training and 2.5k videos for validation. 

\subsection{Evaluation of Sub-task I}
\subsubsection{Implementation Details}
We use the pre-trained MDETR model\footnote{\url{https://github.com/ashkamath/mdetr}} for the object grounding, which is trained on RefCOCO~\cite{yu2016modeling}, RefCOCO+~\cite{yu2016modeling}, RefCOCOg~\cite{mao2016generation}, Flickr30k~\cite{plummer2015flickr30k}, and GQA~\cite{hudson2019gqa} with 1.3 million images-text pairs.
We fine-tune the model for 5 epochs with batch size of 2 on the training set of ActivityNet Entities dataset. Noted that the MDETR model is originally used for image grounding, we extract frames for each video and pair with the video caption to get image-text inputs.
We use EfficientNetb5~\cite{tan2019efficientnet} as the image encoder and RoBERTa~\cite{liu2019roberta} as the text encoder.

\subsubsection{Evaluation Metrics}
We use the localization accuracy~\cite{Zhou_GVD_CVPR19} to evaluate the grounding performance for the sub-task I.
For each object word, only if the IoU of grounded bounding box and the corresponding manual annotation is larger than 0.5, the word is located correctly. The final localization accuracy is computed for each object category and then averaged by the number of unique object categories.

\subsubsection{Experimental Results}
Table~\ref{tab:performance_grounding} shows the grounding performance of the sub-task I on the ActivityNet Entities validation set. Compared with the baseline GVD~\cite{Zhou_GVD_CVPR19}, the localization accuracy is improved to 59.69 by pre-trained MDETR model. After fine-tuning (FMDETR), the performance is further improved to 71.57. In the post processing stage, we fuse 7 models trained by different random seeds (Multi-FMDETR) and it improves the performance to 73.11. 
In addition, combining the model fusion with the co-reference classes unifing (Multi-FMDETR-unify) achieves the best performance of 73.19. The final submitted results for sub-task I on the testing set achieve the localization accuracy of 72.57.

\begin{table}[htbp]
\centering
\caption{Sub-task I: Grounding Performance with ground-truth sentence on the ActivityNet Entities validation set.}
  \begin{tabular}{lc}
    \toprule
     Method & Localization Accuracy(\%)  \\
    \midrule
    GVD & 42.76  \\
    \midrule
     MDETR & 59.69 \\
     FMDETR & 71.57\\
     Multi-FMDETR & 73.11\\
     Multi-FMDETR-unify & \textbf{73.19}\\
    \bottomrule
\end{tabular}
  \label{tab:performance_grounding}
\end{table}

\subsection{Evaluation of Sub-task II}
\subsubsection{Implementation Details}
Our UMPM model consists of four multi-layer transformer encoders, including three independent  encoders  and  a  cross  encoder. The independent encoders contain 1 layer and the cross encoder contains four layers. 
We set the hidden size as 512, attention heads as 8.
The maximum lengths of a video sequence and a caption sentence are set as 70 and 30 respectively.
For the concept prediction, we set the hidden units of LSTM as 512, and pick the top M=15 predicted concepts for each video event. 
We pre-train our UMPM on the ActivityNet Captions~\cite{Heilbron_anet_CVPR15}, TGIF~\cite{Li_TGIF_CVPR16}, MSRVTT~\cite{Xu_MSRVTT_CVPR16}, VATEX~\cite{Wang_VATEX_ICCV19}, Charades~\cite{Sigurdsson_Charades_ECCV16} datasets, with about 770,000 video-caption pairs in total, and fine-tune it only on the in-domain ActivityNet Captions dataset. 
For object grounding, we use Spacy to parse the generated caption sentences. We then keep the nouns and pronouns (except the ``camera'') as object candidates. We finally select the object words from the object  candidates according to the object dictionary and localize them based on MDETR with the same setting as in sub-task I.

\begin{figure*}[htbp]
\centering
\includegraphics[width=\textwidth]{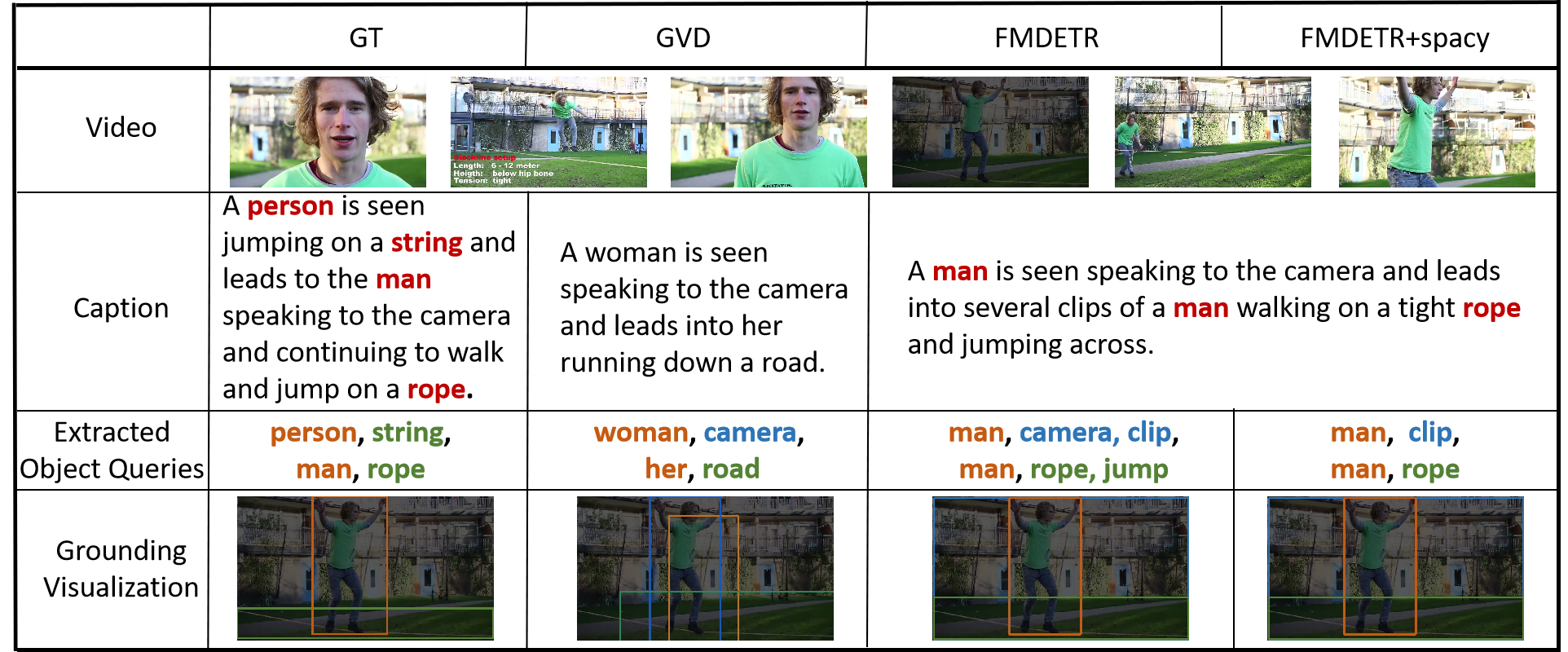}
 \caption{Visualization of grounding performance on generated sentence from different models.}
\label{fig:case_study}
\end{figure*}

\begin{table*}[htbp]
\centering
\caption{Sub-task II: Grounding Performance on generated sentence in ActivityNet Entities validation set.}
  \label{tab:performance_caption_grounding}
  \begin{tabular}{ccccccc}
    \toprule
     Method & P\_all & R\_all & F1\_all & P\_all\_per\_sent & R\_all\_per\_sent & F1\_all\_per\_sent  \\
    \midrule
    GVD &  7.70 & 6.74 & 7.19 & 18.01 & 19.90 & 17.22 \\
    \midrule
     MDETR &11.57 & 18.73&14.31 &18.83 &27.61 &20.26 \\
     FMDETR & 13.71& 22.36 & 16.99 & 21.00 & 34.85& 24.08\\
     FMDETR+Spacy & 13.83 & 22.13 & 17.02 &23.56 & 34.82 & 25.80\\
     Multi-FMDETR+ Spacy & 14.03 & 22.57 & 17.31 & 23.82 & 35.07 & 26.05\\
    \bottomrule
\end{tabular}
\end{table*}




\subsubsection{Evaluation Metrics} 
Sub-task II simultaneously evaluates the captioning performance and grounding performance.
An object word is considered as a correct localization only when it is first correctly predicted and then correctly grounded (IoU \textgreater 0.5).
We use the official metrics, including P\_all, R\_all, F1\_all, P\_all\_per\_sent, R\_all\_per\_sent and F1\_all\_per\_sent. 

\subsubsection{Experimental Results}
Table~\ref{tab:performance_caption_grounding} shows the grounding performance of sub-task II with the generated sentence on the ActivityNet Entities validation set.  Notice that GVD grounds object words generated on its own while MDETR based models ground object words generated by UMPM. We find that without fine-tuning, MDETR provides a much stronger baseline compared with GVD, which demonstrates the advantage of our proposed two-stage manner. 
After fine-tuning with in-domain data, FMDETR further improves. Based on FMDETR, Spacy adjusts the way of selecting object words by filtering verbs and abstract nouns, metric P\_all\_per\_sent improves by more than 2\%. Finally, fusing multiple models helps eliminate contingency and further improves performance to F1\_all\_per\_sent 26.05. The final submitted results for sub-task II on the hidden test set achieve F1\_all\_per\_sent of 24.77.

\subsection{Qualitative Results}
In Figure~\ref{fig:case_study}, we visualize the grounding performance of generated sentence from different models. As for Caption Generation, we highlight the object words in caption with red color. Our UMPM generates captions more accurately as it targets ``man'' and ``rope'' in ground-truth successfully. As for Object Grounding, the grounding box is corresponding to object words with the same color. FMDETR grounds more accurately for word ``man''. When using spacy, words ``jump'' and ``camera'' are recognized as non-object words and won't be localized, which improves the precision of the whole system.

\section{Conclusion}
In this work, we propose a two-stage system for Entities Object Localization. We use a Unified Multi-modal Pre-training Model~(UMPM) to generate video sentence and finetune the pre-trained MDETR~\cite{Kamath_MDETR_arxiv2021} model to ground object words. Our proposed system achieves the state-of-the-art performance on both sub-tasks of Entities Object Localization in the ActivityNet Challenge 2021.
In the future, we will explore the mutual promotion of object localization and caption generation.




\section{Acknowledgments}

This work was supported by National  Natural  Science Foundation of China (Grant No. 61772535 and Grant No. 62072462) and Beijing Natural Science Foundation (Grant No. 4192028).

{\small
\bibliographystyle{ieee_fullname}
\bibliography{main}
}

\end{document}